\documentclass[10pt,twocolumn,letterpaper]{article}

\usepackage[pagenumbers]{wacv} 

\usepackage{graphicx}
\usepackage{amsmath}
\usepackage{amssymb}
\usepackage{booktabs}
\usepackage{times}
\usepackage{epsfig}
\usepackage{subcaption}
\usepackage[table,xcdraw]{xcolor}
\usepackage{multirow}

\usepackage[pagebackref,breaklinks,colorlinks]{hyperref}

\usepackage[capitalize]{cleveref}
\crefname{section}{Sec.}{Secs.}
\Crefname{section}{Section}{Sections}
\Crefname{table}{Table}{Tables}
\crefname{table}{Tab.}{Tabs.}

\def\wacvPaperID{1340} 
\def\confName{WACV}
\def\confYear{2024}

\begin{document}
\title{3D Trajectory Reconstruction of Drones using a Single Camera}

\author{Seobin Hwang, Hanyoung Kim, Chaeyeon Heo, Youkyoung Na, Cheongeun Lee, and Yeongjun Cho
\thanks{Yeongjun Cho is the corresponding author.}\\
Chonnam National University, Gwangju, Republic of Korea\\
{\tt\small cnu.cvl.hsb@gmail.com, codebyhy@gmail.com, cyheo001@gmail.com,} \\
{\tt\small me6zero@jnu.ac.kr, cheongeun02@gmail.com, yj.cho@jnu.ac.kr}
}
\maketitle
\begin{abstract}
Drones have been widely utilized in various fields, but the number of drones being used illegally and for hazardous purposes has increased recently.
To prevent those illegal drones, in this work, we propose a novel framework for reconstructing 3D trajectories of drones using a single camera. 
By leveraging calibrated cameras, we exploit the relationship between 2D and 3D spaces.
We automatically track the drones in 2D images using the drone tracker and estimate their 2D rotations.
By combining the estimated 2D drone positions with their actual length information and camera parameters, we geometrically infer the 3D trajectories of the drones.
To address the lack of public drone datasets, we also create synthetic 2D and 3D drone datasets.
The experimental results show that the proposed methods accurately reconstruct drone trajectories in 3D space, and demonstrate
the potential of our framework for single camera-based surveillance systems.
\end{abstract}

\section{Introduction}
\label{sec:intro}
Recently, drones have been widely utilized in various fields, including security, surveillance, and disaster response~\cite{news01}.
However, drones without permissions for illegal and hazardous purposes have also increased.
For instance, drones equipped with recording devices, such as cameras and storage, can occur security risks.
To address these issues, an anti-drone method that can automatically distinguish illegal drones can be employed.
To this end, estimating 3-dimensional trajectories of drones becomes crucial, because their locations and moving patterns provide significant information e.g., their behavior and legality.

To perform 3D drone trajectory estimation, several approaches~\cite{abdelkrim2008robust,motlagh2019position} have been studied, utilizing various sensors equipped on the object, including Global Positioning System (GPS), Inertial Measurement Unit (IMU), and cameras.
However, in the context of detecting illegal drones, these methods are not suitable, since it is not possible to access the data stored in the drones.
Therefore, we focus on trajectory reconstruction approaches that analyze externally acquired data rather than utilizing internal drone data.
To reconstruct 3D trajectories of objects, numerous studies~\cite{kyriazis2014scalable, nabati2021cftrack, zhang20213d} have proposed methods that utilize various sensors, including Lidar, Radar, and RGB-D cameras. 
They aim to fuse sensor data in order to detect 3D objects and accurately determine their positions.
However, the utilization of multiple sensors can be expensive and impractical.

\begin{figure}[t]
    \centering
    \vspace{10pt} 
    \begin{subfigure}[b]{0.9\linewidth}
        \includegraphics[width=\linewidth]{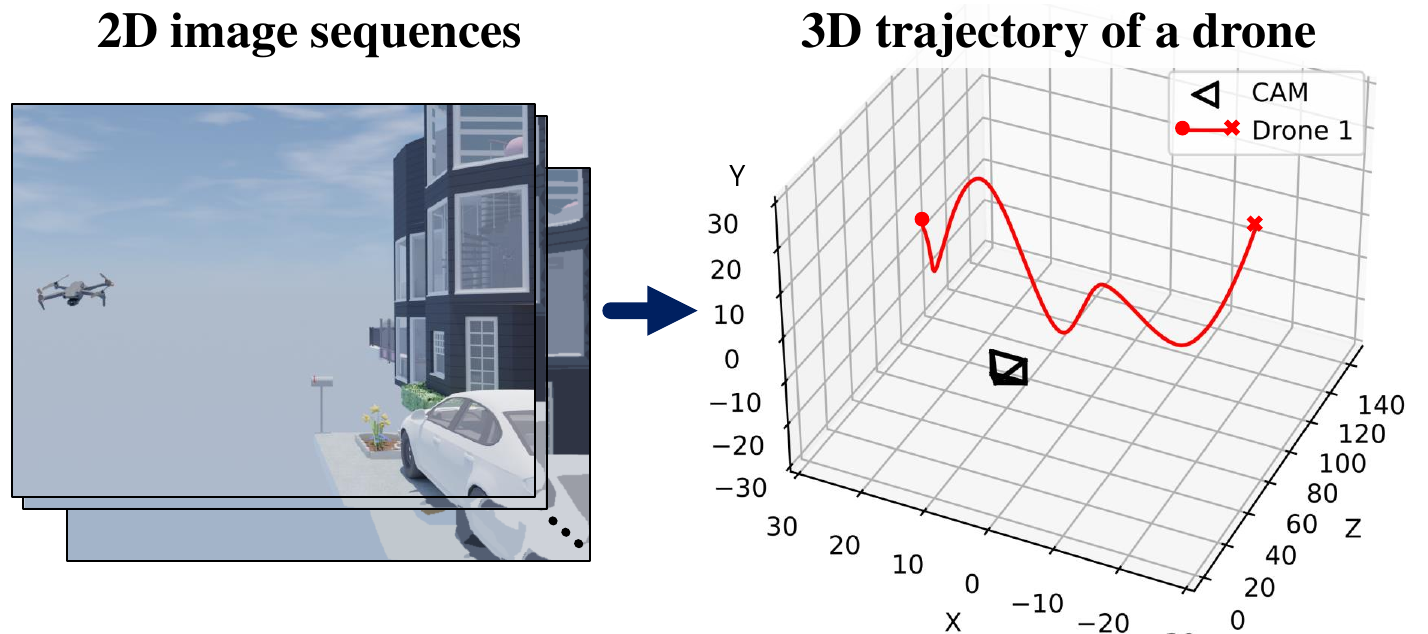}
        \caption{An example of the 3D trajectory reconstruction result}
        \vspace{10pt} 
    \end{subfigure}
    \begin{subfigure}[b]{\linewidth}
        \includegraphics[width=\linewidth]{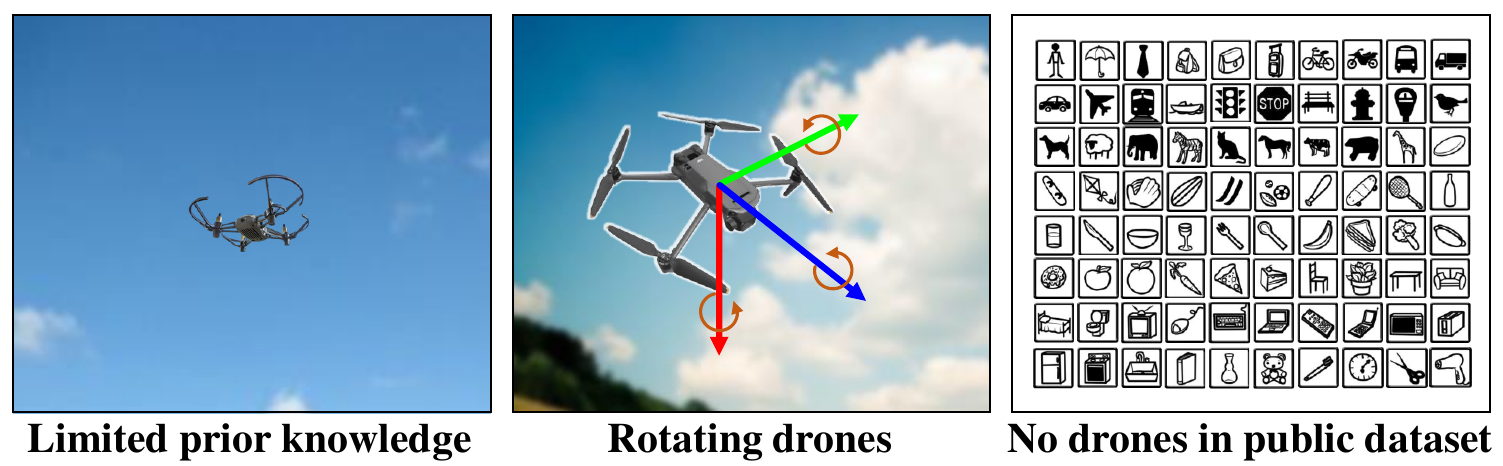}
        \caption{Challenges in 3D trajectory reconstruction of drones}
    \end{subfigure}
    \caption{An example and challenges of 3D trajectory reconstruction of drones using a single camera}
    \label{fig:1}
\end{figure}

In this work, we focus on a surveillance system using a single static camera, since it is cost-effective, and common in urban like a Closed-circuit Television~(CCTV) compared to other multiple sensor-based systems.
Based on these advantages, we propose methods for 3D trajectory reconstruction of drones using a single camera as depicted in Fig.~\ref{fig:1}~(a).
Recently, several methods~\cite{chen2009physics,srinivasan20223d} have attempted to reconstruct the 3D trajectory of objects using a single camera. 
However, they highly rely on prior knowledge, such as ground plane and calibration grids.
Unlike common objects, reconstructing the trajectory of a drone is challenging as shown in Fig.~\ref{fig:1}~(b).
First, drone image scenes are limited prior knowledge e.g., structures of other objects, buildings and ground planes, because drones fly at high altitude.
There is often nothing in the background of the scene except sky and clouds.
Second, drones can rotate freely compared to other common objects.
Third, a drone is not a common object, but a novel object~\cite{horst2016novel} that has not been extensively explored in detection applications. For example, the third figure in Fig.~\ref{fig:1}~(b) shows the object categories provided in MSCOCO~\cite{lin2014microsoft} dataset, but images for drones are missing.

To overcome the limitations, we propose a new framework for 3D trajectory reconstruction of drones using a single camera.
Basically, we utilize calibrated cameras to exploit the relationship between 2D and 3D spaces.
A drone tracker in our framework automatically tracks each drone in the 2D image, and we further find 2D rotations of drones in Sec.~\ref{sec:proposed_2}.
Using estimated 2D drone position and its actual length information from the drone speculation database, we infer the 3D positions of drone as proposed in Sec.~\ref{sec:proposed_3}.
Moreover, we created both 2D and 3D synthetic drone datasets including various scenarios to address the lack of drone datasets in Sec.~\ref{sec:dataset}.

Using our datasets, we validated the effectiveness of proposed methods in Sec.~\ref{sec:exp}.
Despite using a single camera view, our methods accurately reconstructed the 3D trajectories of multiple drones.
The proposed 2D rotation estimation method effectively improved the trajectory reconstruction performance.
In addition, due to our 2D drone image datasets, the performance of drone detector was improved, resulting in improved trajectory reconstruction performance as well.
The results strongly affirm the capability of our framework to accurately reconstruct drone trajectories in 3D space, and demonstrate the potential for applying the framework in single camera surveillance systems.

The main contributions of this work are summarized as follows: 1) A first attempt to reconstruct the 3D trajectory of drones using a single camera, 2) Proposing new methods to overcome the challenges in 3D trajectory reconstruction of drones, 3) Providing new 2D and 3D synthetic datasets of drones.
We hope that this study provides meaningful guidelines to readers who desire to implement a surveillance system for drones in industrial and academic fields.

\section{Related Works}

To localize an object in 3-dimensional space, many studies have been proposed~\cite{kyriazis2014scalable, nabati2021cftrack, zhang20213d} that utilized multiple 3D sensors such as Lidar, Radar, and RGB-D camera. 
Chen \textit{et al.}~\cite{chen2020collaborative} utilized deep neural networks to reconstruct 3D trajectories using spectrum sensing dataset. Nabati \textit{et al.}~\cite{nabati2021cftrack} fused the radar and camera sensors for robust and accurate 3D multi-object tracking.
Meanwhile, many methods based on multiple cameras have been proposed.
For example, Rozantsev \textit{et al.}~\cite{rozantsev2017flight} estimates 6-DOF trajectory of the flying drone using multiple ground cameras. In addition, Li \textit{et al.}~\cite{li2020reconstruction} proposed 3D reconstruction of drone trajectory using multiple unsynchronized cameras with unknown extrinsic camera parameters.
The multiple camera systems can be applied to other applications such as 3D human tracking~\cite{peng2015robust,yao2008multi} and 3d ball tracking~\cite{ren2009tracking, wu2020multi} in many sport scenes.
However, utilizing such sensors or multiple cameras is expensive and also impractical for common surveillance systems (e.g., CCTV).

Geometrically, more than two different camera views are required to reconstruct a 3D trajectory of the object.
Despite of the challenge, several methods have tried to estimate 3D positions of target objects using a single camera.
Rougier \textit{et al.}~\cite{rougier20103d} estimated the 3D trajectory of the human head based on the known human height assumption and calibrated camera.
Similarly, Chen \textit{et al.}~\cite{chen2009physics} employed prior knowledge of a basketball court to reconstruct the 3D trajectory of a basketball.
Recently, Srinivasan \textit{et al.}~\cite{srinivasan20223d} proposed 3D reconstruction of bird flight trajectories with a known calibration grid.
In this way, the aforementioned methods have utilized structural cues of the scenes e.g., ground plane, human heights, court structure, and known grid patterns.
On the other hand, learning-based methods~\cite{garg2019learning,godard2019digging} that directly perform depth estimation using a single camera have been also proposed. 
However, in the case of drone trajectory reconstruction, there is limited prior knowledge available, such as scene and object structure as shown in Fig.~\ref{fig:1}. Therefore, the previous approaches that heavily rely on prior knowledge are ineffective for drone trajectory reconstruction.

\begin{figure*}
	\centering
    \vspace{5pt} 
	\includegraphics[width=1\linewidth]{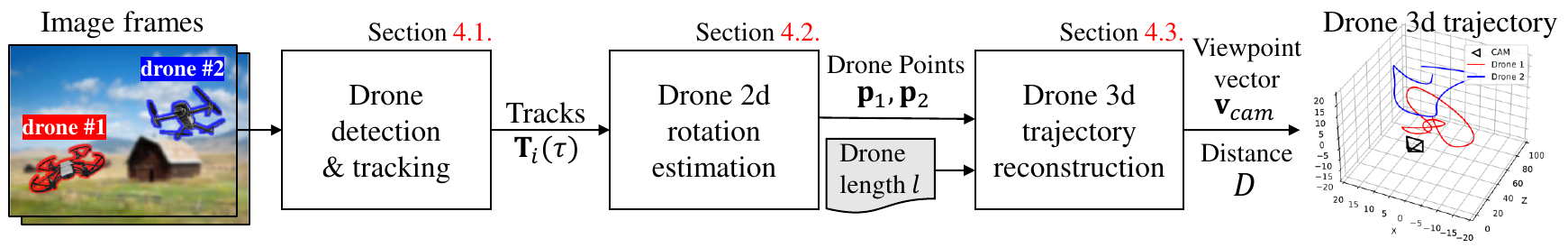}
	\vspace{-15pt}
	\caption{The overall framework for 3D trajectory reconstruction of drones}
	\vspace{-5pt}
	\label{fig:2}
\end{figure*}

\section{Main Idea and Motivation}

According to geometry, at least two different views are required to infer 3D information of objects.
Therefore, reconstructing 3-dimensional trajectories of drones using a single view is very challenging.
Unfortunately, common surveillance systems such as a Closed-circuit Television~(CCTV), generally has a single RGB camera. To make the proposed methods more flexible and applicable, we designed 3D trajectory reconstruction methods that can properly perform using only a single view.
To this end, we focus on the prior knowledge of each drone and camera to estimate the distance between them in 3D space.

In this work, we use calibrated cameras to exploit the relationship between 3D and 2D spaces.
Then we detect each drone in the 2d image, and retrieve actual drone lengths from the drone speculation database.
Based on these cues, we can infer the 3D positions of drone in each frame based on 2D image frames.
The proposed methods can mitigate the reconstruction error with the limited cues. 
Details of the methods are described in Sec.~\ref{sec:proposed}.
In addition, we created both 2d and 3D drone datasets in Sec.~\ref{sec:dataset} for improving and validating the proposed methods.

\section{Proposed Methods}
\label{sec:proposed}
The overall framework for 3D trajectory estimation of drones is illustrated in Fig.~\ref{fig:2}.
It consists of three main parts: 1) drone tracker, 2) drone 2D rotation estimator, 3) 3D drone trajectory re-constructor.
The framework takes consecutive 2d image frames as inputs and generates 3D trajectories of drones in a 3D camera coordinate system as outputs.
First, the drone tracker detects and tracks the drones to estimate their locations, sizes, and identities in 2d image coordinates~(Sec.~\ref{sec:proposed_1}).
Next, we use the information from the tracked drones to estimate 2D rotations. This estimation is performed using Principal Component Analysis (PCA)~\cite{wold1987principal}, and the specific details of this approach are explained in Section \ref{sec:proposed_2}.
Finally, the 3D drone trajectory estimator determines the distance between the camera and drones by leveraging the relationships between them, as well as the actual width lengths of the drones.

\subsection{Drone detection and tracking}
\label{sec:proposed_1}
To estimate 3D drone trajectory, we first estimated 2D trajectories of drones in 2D given image frames.
To this end, we can utilize real-time multiple object trackers~\cite{wojke2017simple,zhang2022bytetrack,zhang2021fairmot}.
Recently, many trackers have adopted tracking-by-detection strategy that associates the detection responses to build object trajectory. Therefore, they essentially include object detector such as YOLO series~\cite{ge2021yolox,redmon2016you} and Faster R-CNN~\cite{ren2015faster} in their frameworks.
The object detection results are represented by $\mathbf{D}_{n}(t)$, where $n$ is an index of detection responses, and $t$ is a frame index. Each detection consists of five-dimensional vectors $\mathbf{D}_{n} = \left(x_{n}, y_{n}, w_{n}, h_{n}, c_{n}\right)$, where $\left(x,y\right)$ is the center position, $\left(w,h\right)$ is the width and height, and $c$ is the class of the drone respectively.
By applying tracking techniques to this information, $\mathbf{D}_{n}$ is refined. As a result, the track is denoted by $\mathbf{T}_{i}(\tau) = \left(\bar{x}_{i}, \bar{y}_{i}, \bar{w}_{i}, \bar{h}_{i}, c_{i}\right)$, where $i$ is an index of the track, $\tau$ is a track lifetime, $\left(\bar{x},\bar{y},\bar{w},\bar{h}\right)$ denotes a $i$th track area estimated by the tracker, and $c_{i}$ is the class of the tracked drone.

Our framework can apply any kind of drone tracker and detector.
However, a large number of drone image dataset is required to train a drone detector.
Unfortunately, drone is considered as a novel object~\cite{horst2016novel} that has not been extensively explored in drone detection applications.
Many public datasets for object detection such as ImageNet~\cite{deng2009imagenet} and MSCOCO~\cite{lin2014microsoft} do not provide images of drones. Although several works~\cite{dronedataset01,coluccia2021drone} for drone image dataset, they provide only a single drone class. 
However, drones come in various sizes and shapes depending on types and brands. Therefore, given that the proposed framework uses actual size of drones, achieving accurate trajectory estimation requires training for each drone class considering unique features.
Considering this issue, we have selected four commonly used drone classes and constructed a 2D drone image dataset in Sec.~\ref{sec:dataset_1}.

\subsection{Drone 2d rotation estimation}
\label{sec:proposed_2}
Drones can freely rotate in the air compared to other common objects. 
Therefore, accurately estimating the rotation of drones is important.
However, the bounding box~(bbox) of a drone only offers a simple rectangular position~($x,y,w,h$) that cannot precisely depict the drone's rotation.
To find the drone rotation in the rectangular bbox, we apply Principal Component Analysis (PCA)~\cite{wold1987principal}.
Based on the tracking result, the estimate position of the drone is defined by $(\bar{x}_{i}, \bar{y}_{i}, \bar{w}_{i}, \bar{h}_{i})$.
Then, a set of pixel coordinates belongs to the foreground drone area is defined by
\begin{equation}
	\mathcal{C} =\{\left(x_k,y_k\right)|1 \leq k \leq  K\},
\end{equation}
where $K$ is the total number of the pixels. We omit the track index $i$ for convenience.
To automatically find the foreground area of the drone, we utilized U2-net~\cite{Qin_2020_PR}.
Then, we calculate the covariance matrix of $\mathcal{C}$ as
{\footnotesize
\begin{equation}
	\mathbf{\Sigma} = \frac{1}{K}\left(\begin{matrix}
	\sum(x_k-m_x)^2  & \sum(x_k-m_x)(y_k-m_y) \\
	\sum(x_k-m_x)(y_k-m_y) & \sum(y_k-m_y)^2 \\
	\end{matrix}\right),
\end{equation}}
where $m_x, m_y$ are means of $x$ and $y$ values, respectively.

\begin{figure}[t]
    \vspace{5pt} 
    \begin{subfigure}[b]{0.495\linewidth}
        \includegraphics[width=\linewidth]{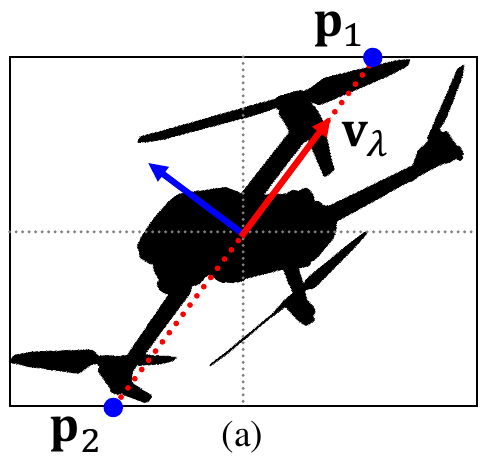}
    \end{subfigure}
    \hfill
    \begin{subfigure}[b]{0.495\linewidth}
        \includegraphics[width=\linewidth]{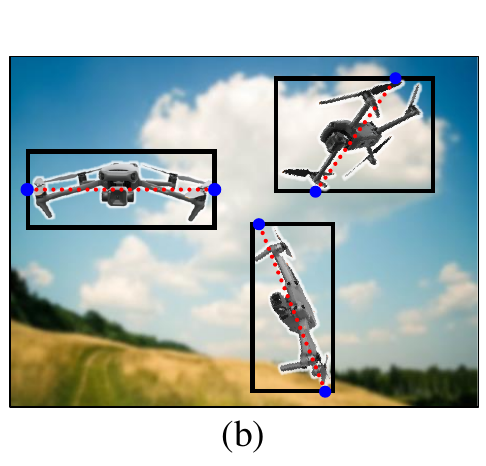}
    \end{subfigure}
    \caption{(a) Description of 2D drone rotation estimation. The black area is the foreground drone area. The arrows represent the eigenvectors of the foreground area. Among these, the red arrow corresponds to the eigenvector $\mathbf{v}_{\lambda}$.  $\mathbf{p}_1, \mathbf{p}_2$ are the intersection points between $(\bar{x}_{i}, \bar{y}_{i}, \bar{w}_{i}, \bar{h}_{i})$ and $\mathbf{v}_{\lambda}$. (b) Examples of 2D drone rotation estimation results.}
	\label{fig:3}
	\vspace{-5pt}
\end{figure}

Based on PCA, two eigenvalues and eigenvectors are calculated from the covariance matrix $\mathbf{\Sigma}$, and we select the eigenvector $\mathbf{v}_{\lambda}$ corresponding to a larger eigenvalue~$\lambda$. 
The eigenvector reflects the rotation of the drone as shown in Fig.~\ref{fig:3}~(a).
Based on the vector, we find two points $(\mathbf{p}_1, \mathbf{p}_2)$ of intersection between the eigenvector and drone position~$(\bar{x}_{i}, \bar{y}_{i}, \bar{w}_{i}, \bar{h}_{i})$. 
To find the intersection points, we define a line that passes through the drone center point~($\bar{x},\bar{y}$), with its direction determined by the eigenvector $\mathbf{v}_{\lambda}$. As a result, we can approximately estimate the 2D rotations of drones regardless of their poses as shown in Fig.~\ref{fig:3}~(b). This aids in enhancing the overall accuracy of the drone's 3D trajectory reconstruction.

\subsection{Drone 3D trajectory reconstruction}
\label{sec:proposed_3}
To reconstruct the 3D trajectory of the drone, in this section, we exploit the prior knowledge of each drone such as its specification information.
We assume that the principle line $\overline{\mathbf{p}_1\mathbf{p}_2}$ of the drone corresponds to the the longest side of the drone. In general, the width length of the drone is longer than its other sides e.g., height and depth.
Based on this assumption, we retrieve an actual drone's width length~$l$ from the drone specification database when the drone detector predicts the class of the drone as $c$.
Then, we have two end points~($\mathbf{p}_1,\mathbf{p}_2$) of the drone in 2d image, and its real 3D length $l$.

Using those cues (i.e., $\mathbf{p}_1,\mathbf{p}_2, l$) of the drone, we can estimate 3D position of the drone. To this end, we calibrate the camera to find the relationship between 2d image coordinate~($u,v$) and 3D world coordinate systems~($X,Y,Z$) as
\begin{equation}
	\begin{bmatrix}u \\ v \\ 1 \\ \end{bmatrix} = \mathbf{K} [\mathbf{R}|\mathbf{t}] \begin{bmatrix}X \\ Y \\ Z \\ 1\\ \end{bmatrix},
\end{equation}
where $\mathbf{K}^{3\times3}$ matrix denotes camera intrinsic parameters, $\mathbf{R}^{3\times3},\mathbf{t}^{3\times1}$ denote camera extrinsic parameters~(rotation, translation).
By calculating $[\mathbf{R}|\mathbf{t}] \cdot [X,Y,Z,1]^{\top}$, the world coordinate system is transformed into a camera coordinate system as
\begin{equation}
	\begin{bmatrix}u \\ v \\ 1 \\ \end{bmatrix} = \textbf{K} \begin{bmatrix}X_{c} \\ Y_{c} \\ Z_{c} \\ \end{bmatrix}.
\end{equation}

\begin{figure}[t]
    \vspace{5pt} 
	\begin{center}
		\includegraphics[width=1\linewidth]{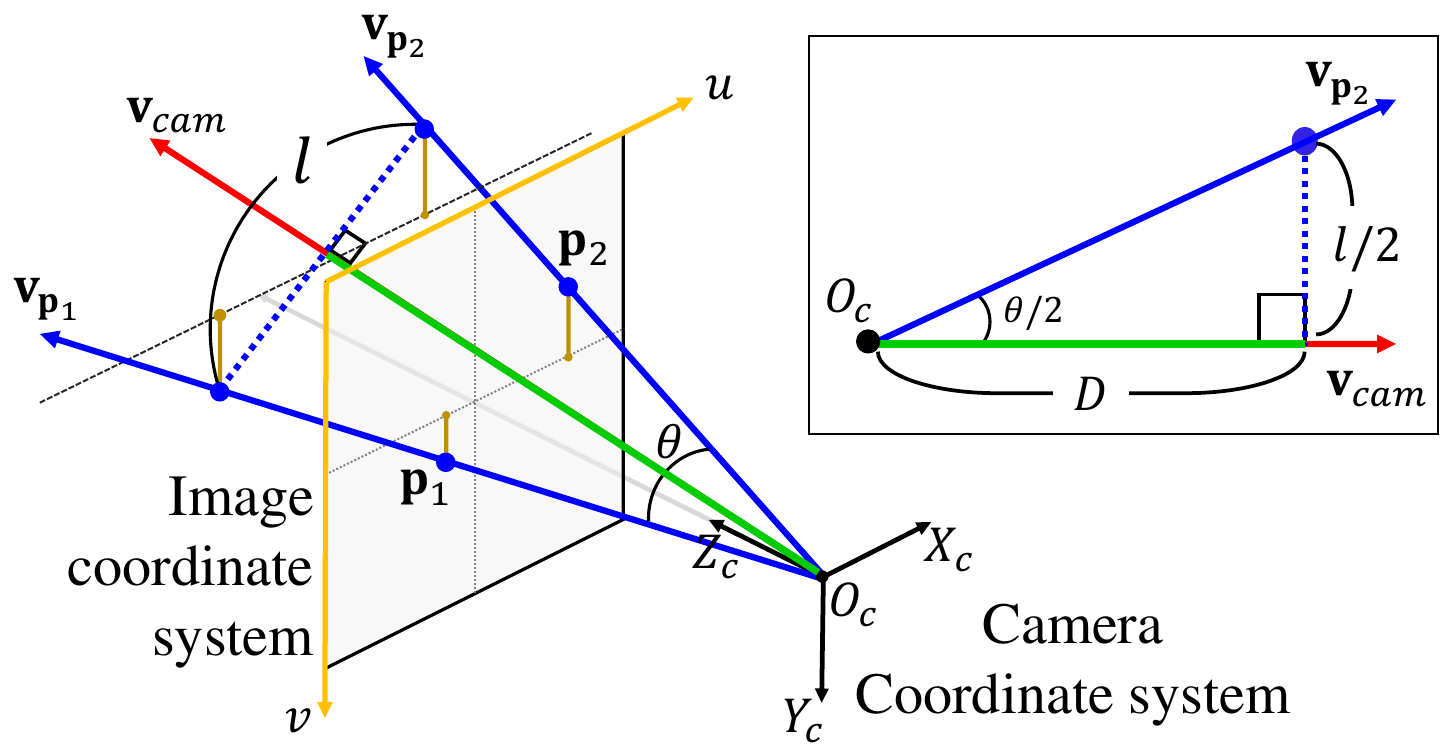}
	\end{center}
	\caption{Geometric relationship between camera and image coordinate systems. The green line is the distance $D$ between the center of drone and camera. Best viewed in color.}
	\label{fig:4}
\end{figure}

In the camera coordinate system, the origin $O_c=[0,0,0]$ represents the center of the camera.
Since the $\mathbf{K}$ matrix is invertible, we can have a linear transformation of the 2d image coordinate to the 3D camera coordinate system by
\begin{equation}
	\begin{bmatrix} X_{c} \\ Y_{c} \\ Z_{c} \\ \end{bmatrix} = \textbf{K}^{-1} \begin{bmatrix} u \\ v \\ 1 \\ \end{bmatrix}.
\label{eq:5}
\end{equation}
According to Eq.~\ref{eq:5}, we can transform the points~($\mathbf{p}_1,\mathbf{p}_2$) in the 2d image coordinates system into 3D vectors in the camera coordinate system by calculating
\begin{equation}
	\mathbf{v}_{\mathbf{p}_1} = \mathbf{K}^{-1}\begin{bmatrix} \mathbf{p}_1  \\ 1 \end{bmatrix}, \quad
	\mathbf{v}_{\mathbf{p}_2} = \mathbf{K}^{-1}\begin{bmatrix} \mathbf{p}_2  \\ 1 \end{bmatrix}.
\end{equation}
Note that the vectors $\mathbf{v}_{\mathbf{p}_1}$ and $\mathbf{v}_{\mathbf{p}_2}$ are heading to the points ($\mathbf{p}_1,\mathbf{p}_2$) from the origin~($O_c$).
Additionally, a camera viewpoint vector to the center of the drone is defined by $\mathbf{v}_{cam}=\frac{\mathbf{v}_{\mathbf{p}_1}+\mathbf{v}_{\mathbf{p}_2}}{2}$.
Figure.~\ref{fig:4} visualizes both image and camera coordinates with the 2D image points~($\mathbf{p}_1,\mathbf{p}_2$) and vectors~($\mathbf{v}_{\mathbf{p}_1},\mathbf{v}_{\mathbf{p}_2},\mathbf{v}_{cam}$).

Assuming that the vector $\overrightarrow{\mathbf{v}_{\mathbf{p}_1}\mathbf{v}_{\mathbf{p}_2}}$ is orthogonal to a camera view point vector $\mathbf{v}_{cam}$, then we can calculate the distance between   the drone and camera center by following steps.
First, compute a vector angle~$\theta$ between $\mathbf{v}_{\mathbf{p}_1}$ and $\mathbf{v}_{\mathbf{p}_2}$ by
\begin{equation}
	\theta = cos^{-1}\frac{\mathbf{v}_{\mathbf{p}_1}\cdot \mathbf{v}_{\mathbf{p}_2}}{\left\|\mathbf{v}_{\mathbf{p}_1} \right\| \left\|\mathbf{v}_{\mathbf{p}_2} \right\|}.
\end{equation}
The distance between two vectors is known as $l$, which is an actual drone length. Based on the $\theta$ and $l$ values, the distance between the camera and the drone can be calculated by the following trigonometric transformation, by
\begin{equation}
	D = \frac{l}{2 \tan\frac{\theta}{2}}.
\end{equation}
Then, we can determine the 3D position of the drone according to the distance $D$ and camera view point vector $\mathbf{v}_{cam}$.
3D position estimation for drones can be conducted across multiple image frames to build the 3D trajectory.
However, the reconstructed 3D trajectory using a single camera may contain noise and error. 
To mitigate them, we can apply an average filter to the initial trajectory.
By smoothing out variations in the trajectory, the average filter helps enhance the accuracy of the drone's 3D position estimation.

\section{Datasets}
\label{sec:dataset}
\subsection{\texttt{\textbf{2Drone(on+aug)}} -- 2D drone image set}
\label{sec:dataset_1}

Many studies have proposed the methods to handle novel object problems~\cite{horst2016novel,zhou2021image}.
A drone is also a novel object that has not been extensively explored in object detection applications.
Public object detection datasets e.g., ImageNet~\cite{deng2009imagenet} and MSCOCO~\cite{lin2014microsoft} do not provide drone images for training the detector. 
Several works~\cite{dronedataset01,coluccia2021drone} built drone image datasets, but they only provide a single drone class.
Considering the issues, we created a new 2D drone image dataset called \texttt{2Drone} for training drone detectors.
It provides multi-class drone models~(Air2S, Mavic3, Mini3Pro, and Tello) with diverse poses and backgrounds.
The drones developed by DJI Technology are popular and widely utilized.

\begin{table}[]
	\centering
	\small
	\renewcommand{\arraystretch}{1.2}
	\begin{tabular}{c|c|c}
		\hline
		\noalign{\hrule height 1pt}
		\rowcolor[HTML]{EFEFEF} 
		Category               & Name   & Numbers of source images \\ \hline
		\multirow{4}{*}{Drone} & Airs2S & 25             \\ \cline{2-3} 
		& Mavic3 & 32             \\ \cline{2-3} 
		& Mini3  & 39             \\ \cline{2-3} 
		& Tello  & 41             \\ \hline
		Background             & --      & 100           \\ \hline\noalign{\hrule height 1pt}
	\end{tabular}
	\caption{Summary of source images for 2D drone image dataset}
	\vspace{5pt}
	\label{tab:01}
\end{table}

\begin{figure}[t]
    \centering
    \begin{minipage}[b]{0.45\columnwidth}
        \centering
        \includegraphics[height=0.65\columnwidth]{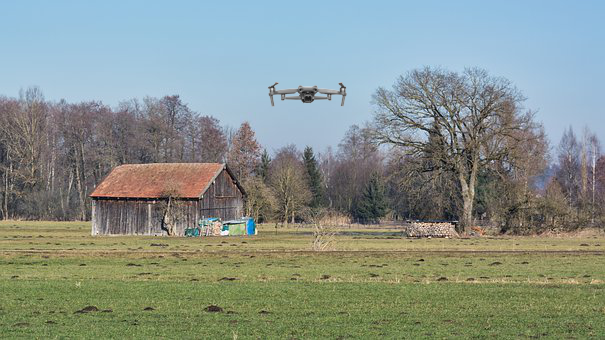}
        \caption*{(a) Airs2S}
    \end{minipage}\hfill
    \begin{minipage}[b]{0.45\columnwidth}
        \centering
        \includegraphics[height=0.65\columnwidth]{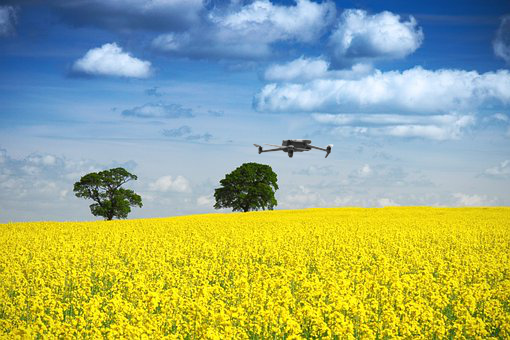}
        \caption*{(b) Mavic3}
    \end{minipage}
    
    \vspace{10pt}
    
    \begin{minipage}[b]{0.45\columnwidth}
        \centering
        \includegraphics[height=0.65\columnwidth]{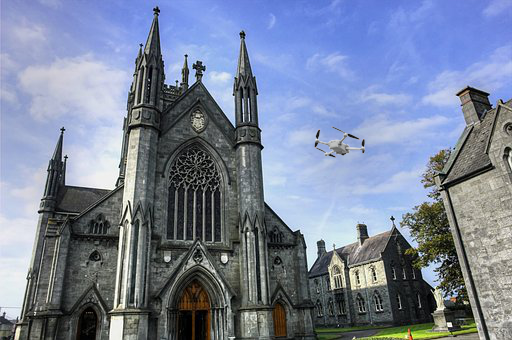}
        \caption*{(c) Mini3}
    \end{minipage}\hfill
    \begin{minipage}[b]{0.45\columnwidth}
        \centering
        \includegraphics[height=0.65\columnwidth]{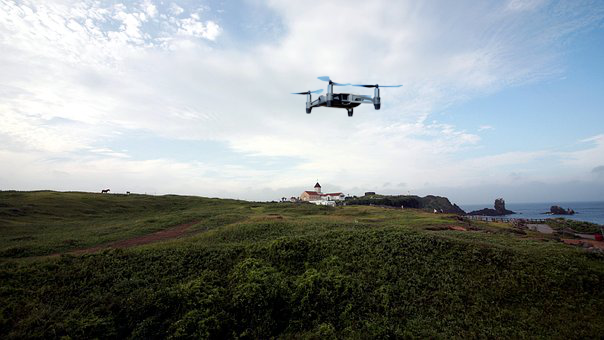}
        \caption*{(d) Tello}
    \end{minipage}
    \hspace{14pt}
    \caption{Examples of synthesized 2D drone images}
    \label{fig:05}
    \vspace{-10pt}
\end{figure}

We first collected 1,107 numbers of real drone images of the four models available online and named this drone image set as \texttt{2Drone(on)}.
However, the volume of the dataset is not sufficient to train a robust and accurate drone detector, and we realized that the public images of that drones are limited.
As the work~\cite{block2022image} synthesized objects and background images to make an object detection dataset, we synthesized drone source images with background images to augment more 2D drone image dataset.
To this end, we collected source images of the four drones with different directions. In addition, inspired by a work~\cite{xiao2020noise}, we also collected various royalty-free background images as summarized in Table~\ref{tab:01}.

Then, we randomly select drone and background source images, and mix them as shown in Fig.~\ref{fig:05}. 
Drones are typically found in the upper airspace. 
Therefore, the positions of drones were determined randomly around the upper part of the background images.
The total number of the images is 11,434, and the number of each drone in the images is as follows -- Airs2S: 2,085, Mavic3: 2,668, Mini3: 3,245, Tello: 3,436. We call the dataset \texttt{2Drone(aug)}.
To sum up, our dataset named \texttt{2Drone(on+aug)} consists of \texttt{2Drone(on)} and \texttt{2Drone(aug)}.
It involves 12,541 images and their ground-truth bounding boxes. It is available on \url{https://will_be_available}.

\begin{figure*}[t]
	\centering
	\begin{minipage}[b]{0.63\columnwidth}
		\centering
		\includegraphics[width=\linewidth]{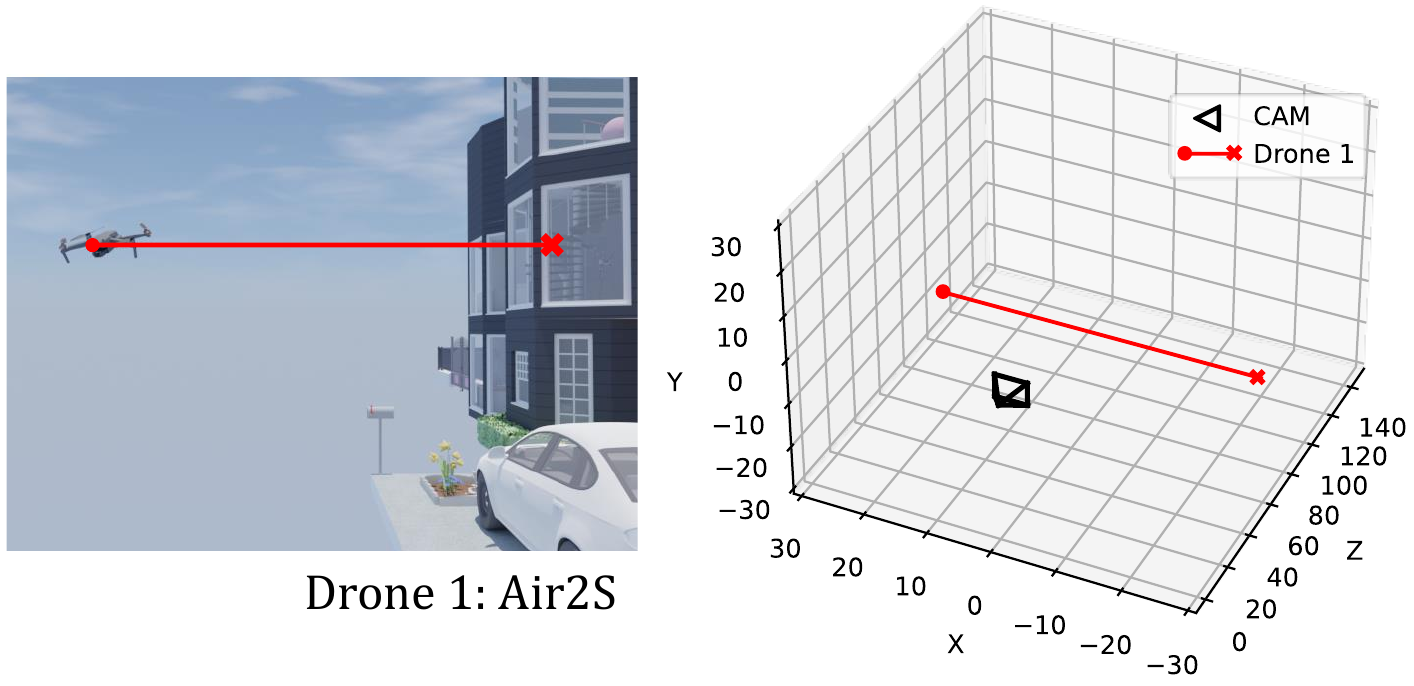}
		\caption*{Sequence \#01}
	\end{minipage}\hspace{15pt}
	\begin{minipage}[b]{0.63\columnwidth}
		\centering
		\includegraphics[width=\linewidth]{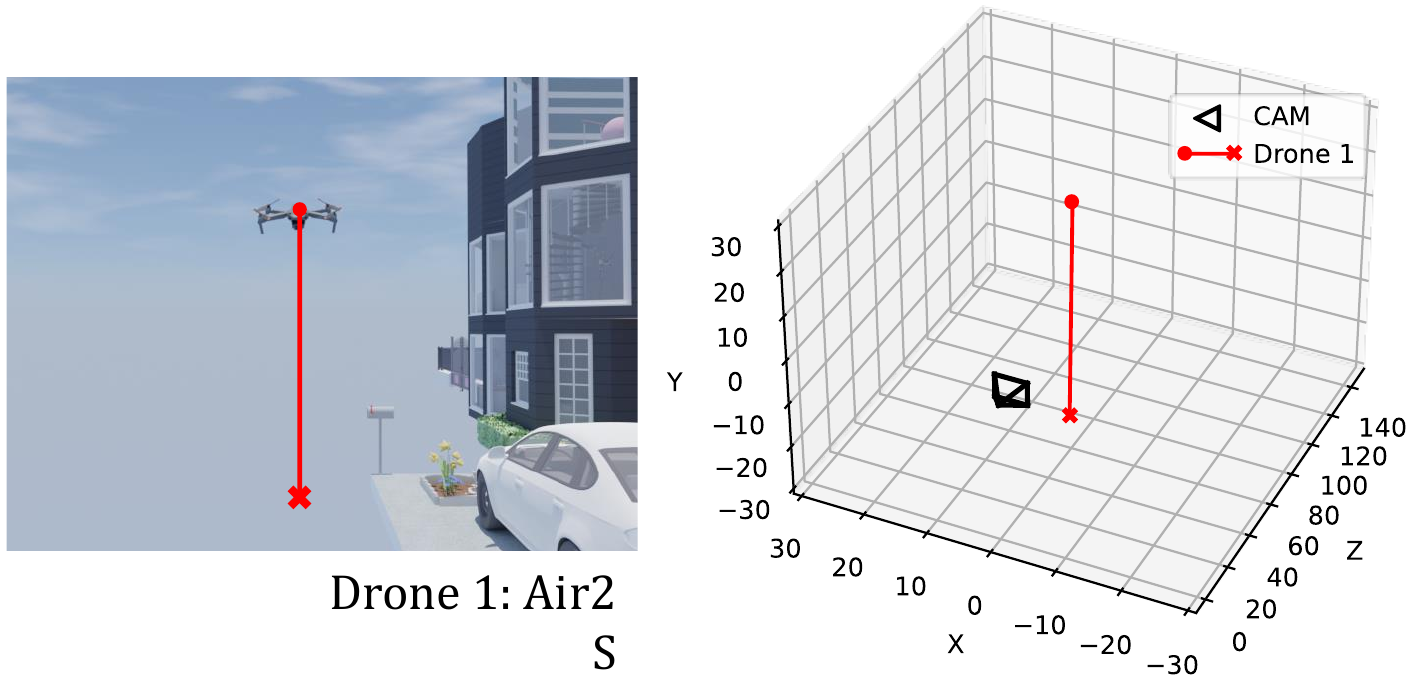}
		\caption*{Sequence \#02}
	\end{minipage}\hspace{15pt}
	\begin{minipage}[b]{0.63\columnwidth}
		\centering
		\includegraphics[width=\linewidth]{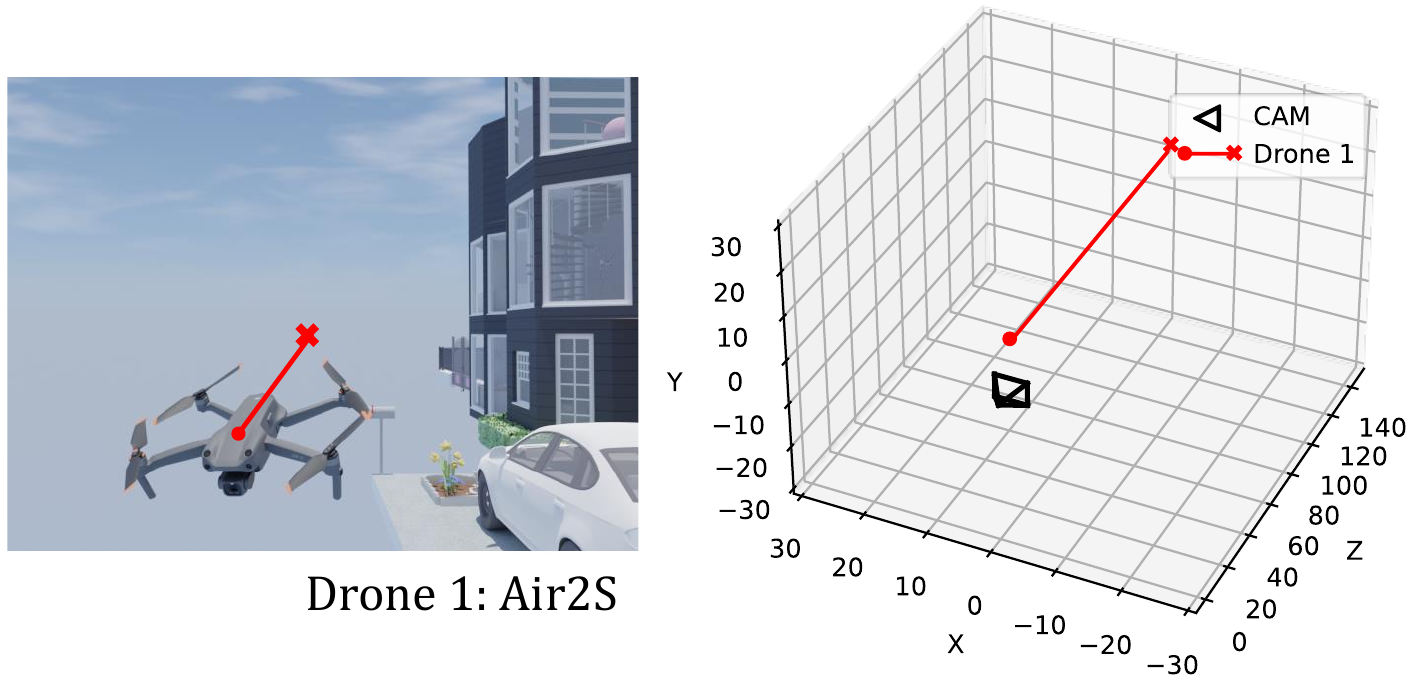}
		\caption*{Sequence \#03}
	\end{minipage}\vspace{10pt}
	\begin{minipage}[b]{0.63\columnwidth}
		\centering
		\includegraphics[width=\linewidth]{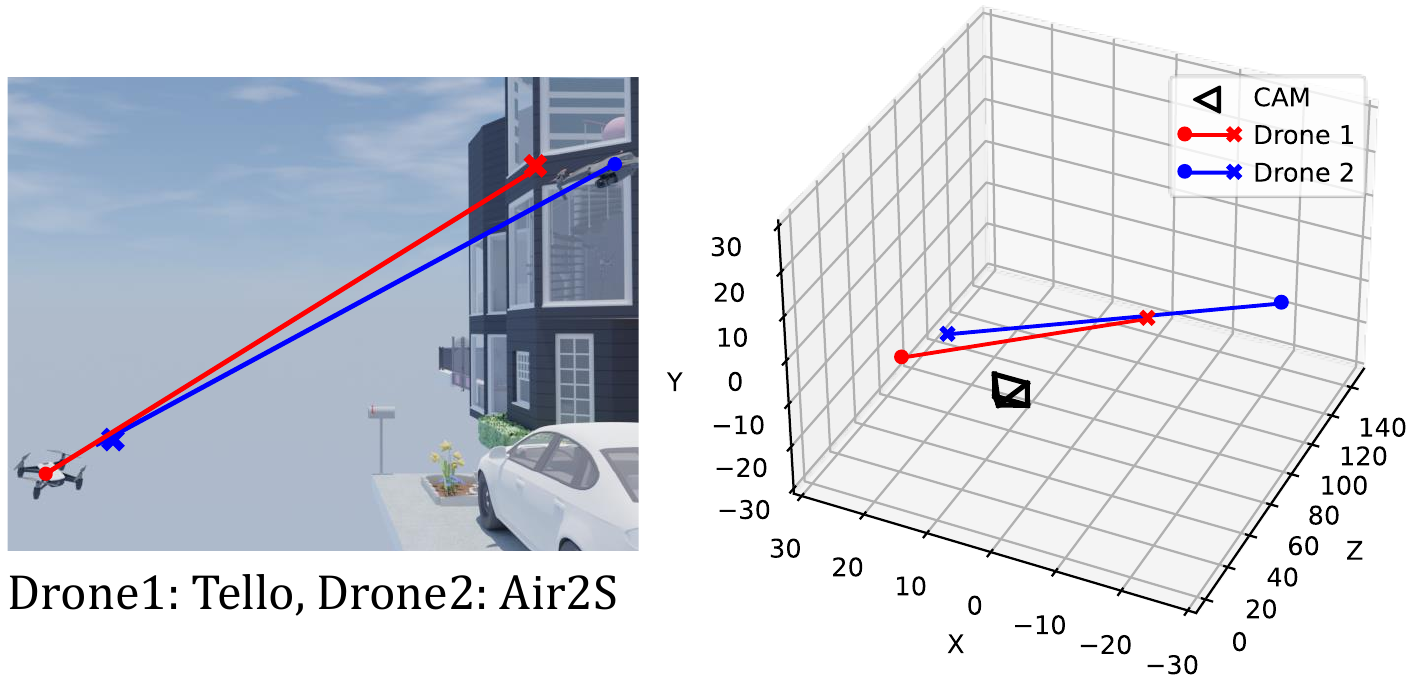}
		\caption*{Sequence \#04}
	\end{minipage}\hspace{15pt}
	\begin{minipage}[b]{0.63\columnwidth}
		\centering
		\includegraphics[width=\linewidth]{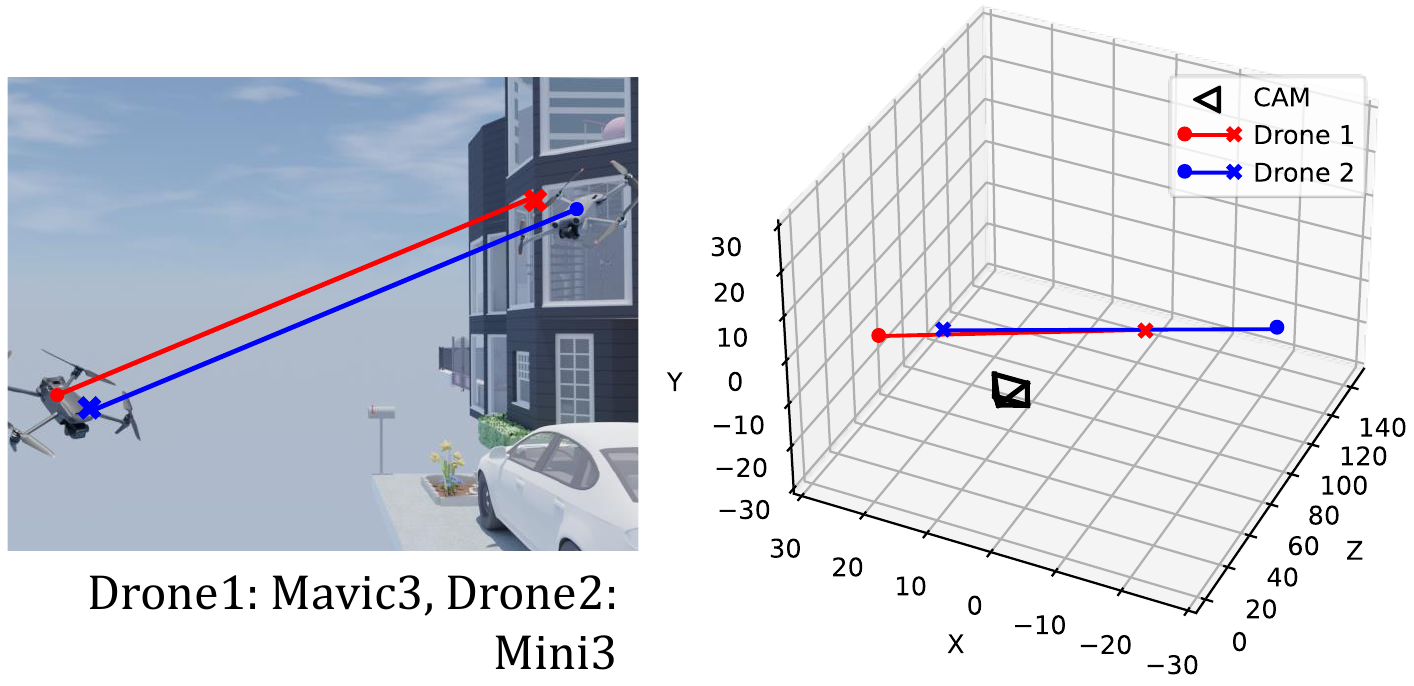}
		\caption*{Sequence \#05}
	\end{minipage}\hspace{15pt}
	\begin{minipage}[b]{0.63\columnwidth}
		\centering
		\includegraphics[width=\linewidth]{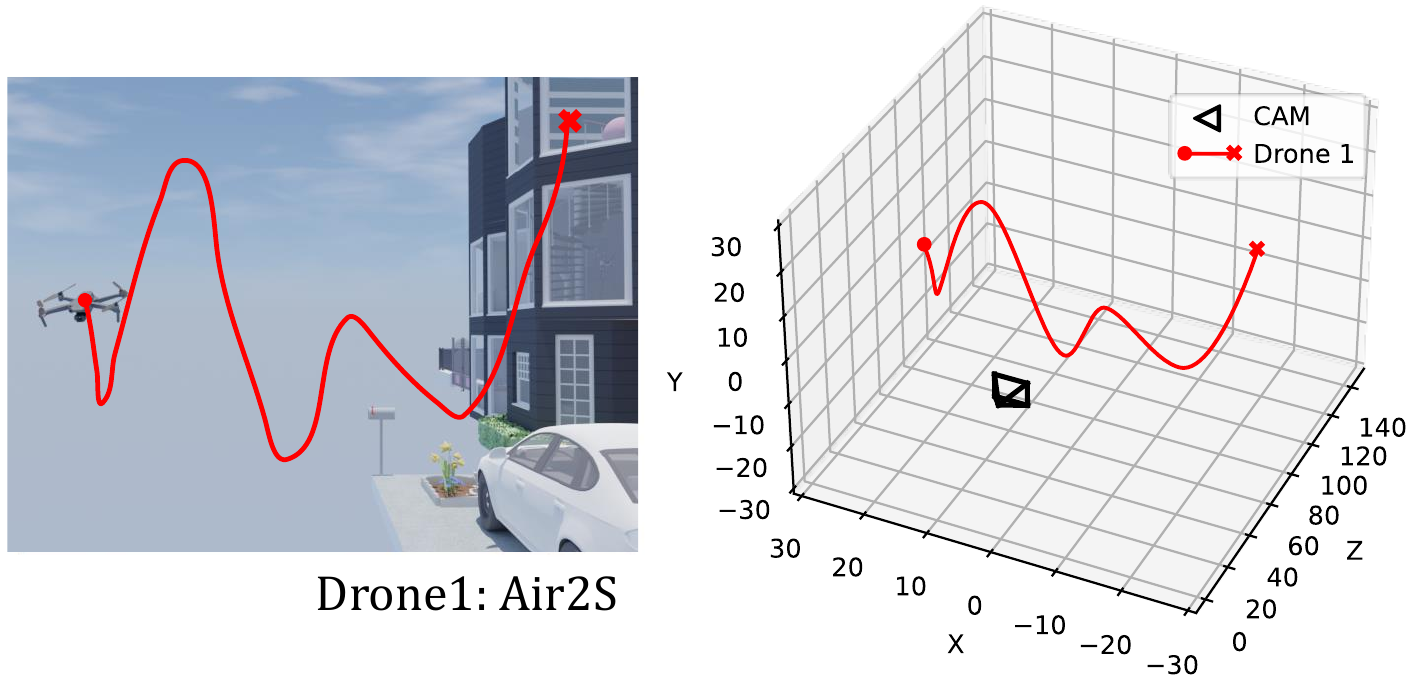}
		\caption*{Sequence \#06}
	\end{minipage}\vspace{10pt}
	\begin{minipage}[b]{0.63\columnwidth}
		\centering
		\includegraphics[width=\linewidth]{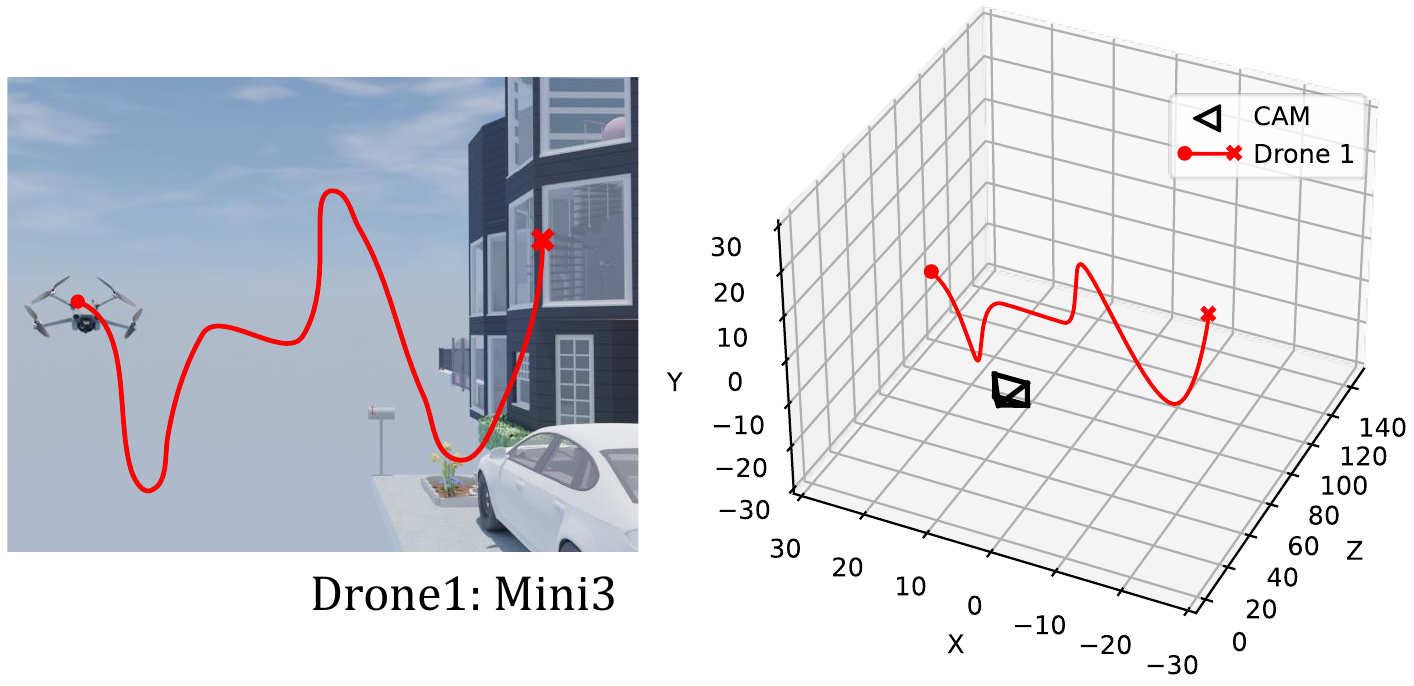}
		\caption*{Sequence \#07}
	\end{minipage}\hspace{15pt}
	\begin{minipage}[b]{0.63\columnwidth}
		\centering
		\includegraphics[width=\linewidth]{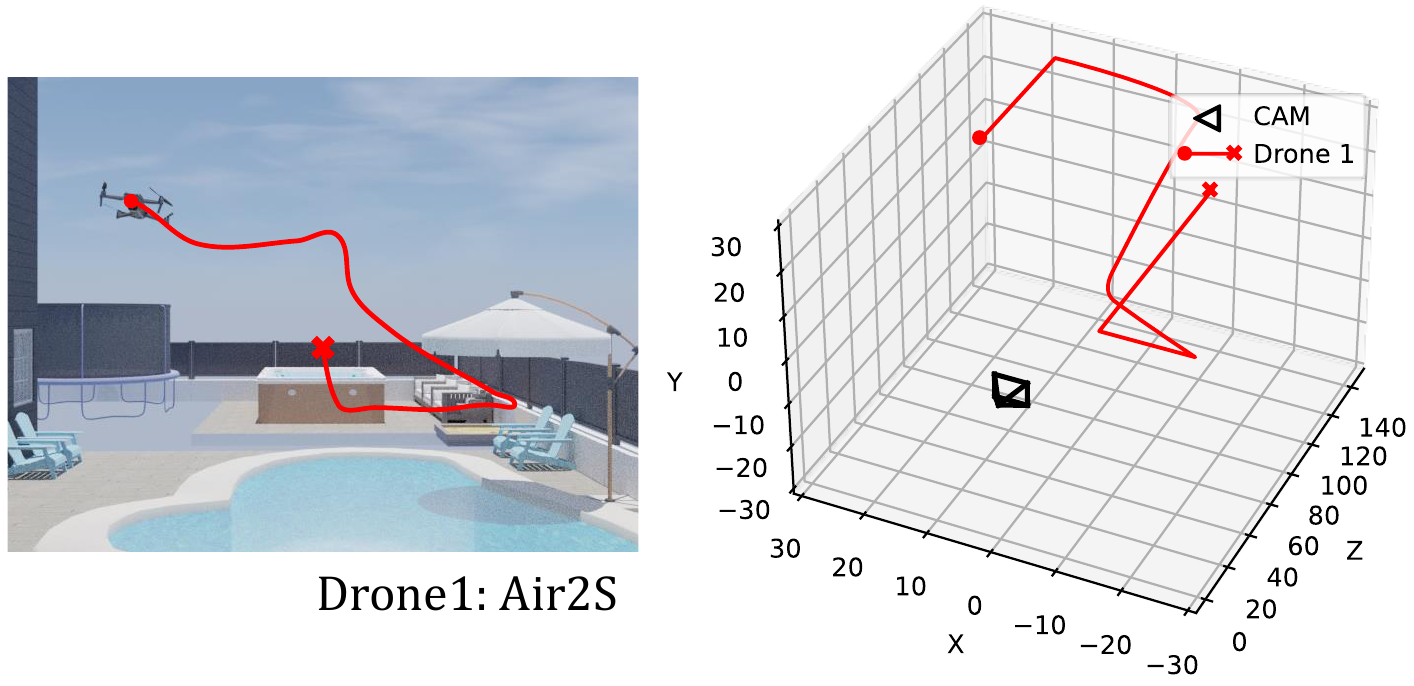}
		\caption*{Sequence \#08}
	\end{minipage}\hspace{15pt}
	\begin{minipage}[b]{0.63\columnwidth}
		\centering
		\includegraphics[width=\linewidth]{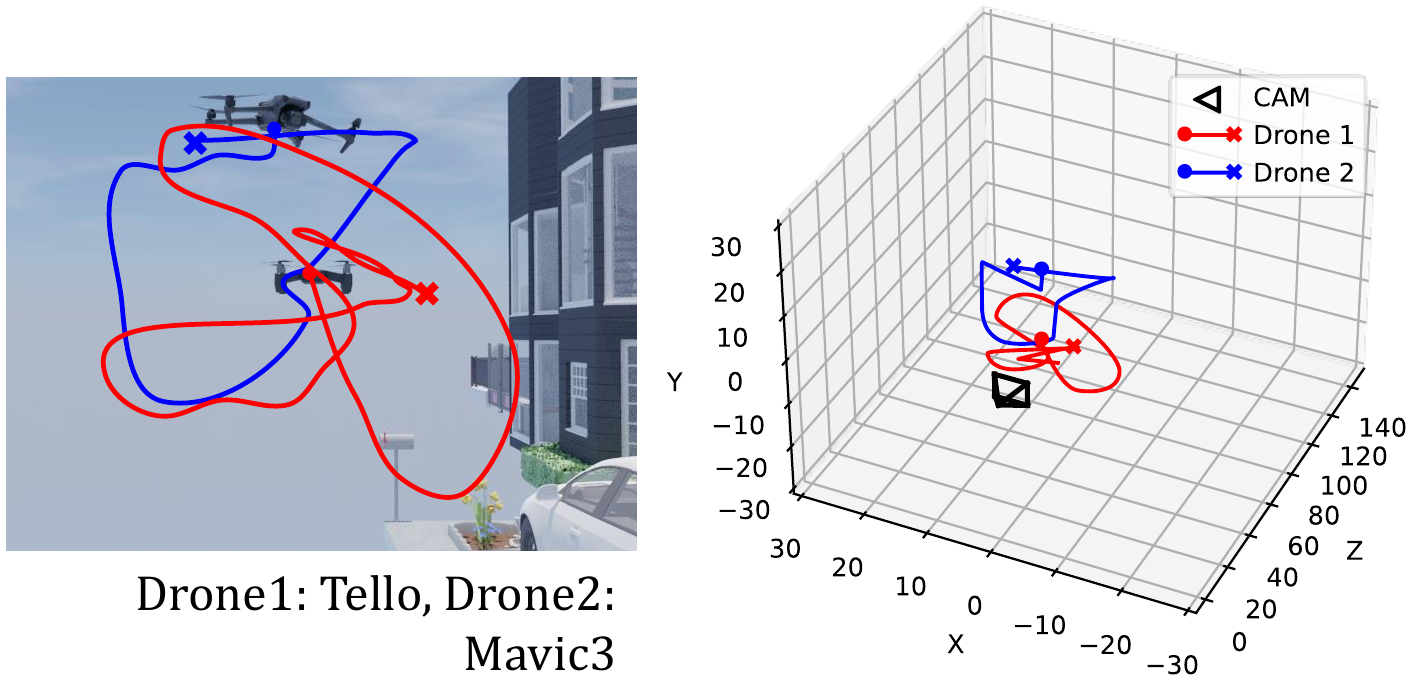}
		\caption*{Sequence \#09}
	\end{minipage}
	\caption{Examples of image frames and corresponding 3D drone trajectories of the 3D drone synthetic dataset. $\bullet$ and $\times$ symbols denote the start and end of the drone trajectory, respectively.}
	\label{fig:06}
\end{figure*}

\begin{table*}[t]
	\centering
	\small
	\renewcommand{\arraystretch}{1.2}
	\begin{tabular}{r||c|c|c|c|c|c|c|c|c}
		\hline\noalign{\hrule height 1pt}
		\rowcolor[HTML]{EFEFEF} 
		Sequences     & Seq. \#01  & Seq. \#02  & Seq. \#03  & Seq. \#04 & Seq. \#05& Seq. \#06& Seq. \#07 &Seq. \#08 &Seq. \#09 \\ \hline
		Drones        & Air2S    & Air2S    & Air2S    &\begin{tabular}[c]{@{}c@{}}Air2S\\ Tello\end{tabular}&\begin{tabular}[c]{@{}c@{}}Mavic3\\ Mini3\end{tabular}&Air2S & Mini3& Air2S & \begin{tabular}[c]{@{}c@{}}Tello\\ Mavic3\end{tabular}\\ \hline
		Frames        &   180    &   180    &    180   &     180 &   180  & 300 &  300 & 810  &  300 \\ \hline
		Translation     &   $x$      &  $y$       & $z$        & $x,y,z$   &  $x,y,z$ &$x,y,z$&$x,y,z$&$x,y,z$ & $x,y,z$\\ \hline
		Rotation      &   None      & None         &None     & None      &  None   &  None &  None   &$\checkmark$&$\checkmark$\\ \hline
		Motion        &  Linear    & Linear     &   Linear   & Linear    &   Linear &Non-linear&Non-linear&Non-linear&Non-linear\\ \hline\noalign{\hrule height 1pt}
	\end{tabular}
	\vspace{5pt}
	\caption{Properties of scenarios in proposed 3d drone dataset. Translation denotes the direction of the drone's movements. Rotation denotes the changes of drone's pose angle. Motion (linear) and Motion (non-linear) mean the drone movements with constant and variable velocity.}
	\label{tab:02}
\end{table*}

\subsection{\texttt{\textbf{Syn3Drone}} -- Synthetic 3D drone trajectory set}
\label{sec:dataset_2}
To validate the proposed methods, we require a 3D drone trajectory dataset called \texttt{Syn3Drone} that provides both 2D image frames and ground-truth drone 3D positions. As we know, there is no public dataset for the 3D drone tracking.
In this work, we newly built a synthetic 3D drone trajectory dataset by utilizing an open-source 3D computer graphics software tool~(blender). 
We collected 3D models of the four drones~(Airs2S, Mavic3, Mini3, Tello), and constructed two different 3D background scenes.
Then, we rendered 9 different drone scenarios as depicted in Fig.~\ref{fig:06}.
The properties of sequences in the dataset are summarized in Tab.~\ref{tab:02}.
The image frame resolution of each sequence is $640\times480$, and the frame rate is $30$.
We set the drones considering their actual sizes. The width, depth and height lengths~(mm) of drones are as follows -- Air2S {\small(W:253.0, D:183.0, H:77.0)}, Mavic3 {\small(W:347.5, D:283.0, H:107.7)}, Mini3 {\small(W:245.0, D:171.0, H:62.0)}, and Tello {\small(W:176.3, D:98.0, H:41.0)}.

The features of the sequences are as follows.
First, Seq.~\#01 -- Seq.~\#03 are the straightforward scenario including a single drone exhibiting linear motion along a single axis ($x$, $y$, or $z$) without any rotation.
To consider multiple drones, Seq.~\#04 and Seq.~\#05 exhibit two drone crossing scenario with a linear drone motion without rotation.
Seq.~\#06 and Seq.~\#07 contain a single drone with nonlinear motion.
Furthermore, Seq.~\#08 and Seq.~\#09 are the complex scenario that include a rotating drone with the nonlinear motion.
Based on various drone scenarios, we expect to be able to verify the method in various aspects.
All the rendered 2D image sequences and 3D drone trajectory information are available on \url{https://will_be_available}.

\section{Experimental Results}
\label{sec:exp}

\subsection{Settings and evaluation metrics}

For the multi-object tracker, we utilized byteTrack~\cite{zhang2022bytetrack} that can perform real-time tracking.
An object detector for the tracker, we trained a YOLOv5x~\cite{yolov5} model using NVIDIA RTX 3060 with the following hyper-parameter settings: 100 epochs, a batch size of 4, and a learning rate of 0.0001. 
Using the proposed dataset in Sec.~\ref{sec:dataset_1}, we evaluated the drone detector with various training \& test scenario.
To evaluate the accuracy of multi-object trackers, we measure the Multi-Object Tracking Accuracy~(MOTA)~\cite{bernardin2008evaluating} defined by $MOTA = 1 - \frac{ \sum_t FN_t + FP_t + IDs_t }{\sum_t g_t },$ where where $FN_t$, $FP_t$, $IDs_t$, and $g_t$ are the number of misses (false negatives), false positives, identity switches, and objects, respectively.
To evaluate performance of 3D trajectory estimation, we measure two commonly used metrics: Mean Absolute Error~(MAE) and Root Mean Squared Error~(RMSE).
MAE calculates the absolute differences between the predicted and actual values, and measures an average summation of the differences. 
It provides an average measure of the absolute deviation between the predicted and actual values and it is not sensitive to outliers, as it treats all differences equally~\cite{chai2014root}.
RMSE also calculates the differences between the predicted and actual values, but it further squares the differences for the average of the squared differences.
Small value indicates high performance for the both metrics.

\begin{table}[t]
	\centering
	\setlength\tabcolsep{4.5pt}
	\renewcommand{\arraystretch}{1.2}
	\small
	\begin{tabular}{r||c|c|c|c}
		\hline\noalign{\hrule height 1pt}
		\rowcolor[HTML]{EFEFEF}
		Training Dataset  & MOTA$(\uparrow)$ & FN$(\downarrow)$ & FP$(\downarrow)$ & IDs$(\downarrow)$  \\ \hline
		\texttt{2Drone(on)}  & 75.3 &693 & \textbf{58}  &5\\ \hline
		\texttt{2Drone(aug)} & 80.0 &393 &214 &6\\ \hline
		\texttt{2Drone(on+aug)}  & \textbf{81.0} & \textbf{343} &236 & \textbf{3}\\ \hline\noalign{\hrule height 1pt}
	\end{tabular}
	\vspace{5pt}
	\caption{Multi-drone tracking results on \texttt{Syn3Drone} dataset according to different training datasets for the drone detector.}
	\vspace{-5pt}
	\label{tab:03}
\end{table}

\begin{figure}[t]
	\centering
	\begin{minipage}[b]{0.315\linewidth}
		\centering
		\includegraphics[width=\linewidth]{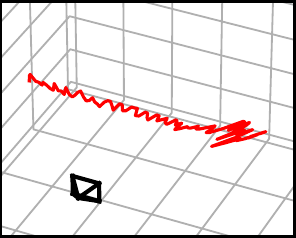}
		\caption*{Initial estimation}
	\end{minipage}
	\begin{minipage}[b]{0.315\linewidth}
		\centering
		\includegraphics[width=\linewidth]{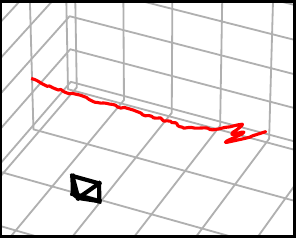}
		\caption*{Window size: 5}
	\end{minipage}
	\begin{minipage}[b]{0.315\linewidth}
		\centering
		\includegraphics[width=\linewidth]{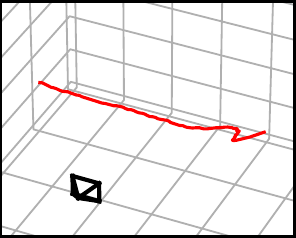}
		\caption*{Window size: 10}
	\end{minipage}
	\caption{Trajectory smoothing results according to window sizes}
	\label{fig:7}
\end{figure}

\subsection{Multi-drone tracking results}
For the multi-drone tracker, we trained various drone detectors according to our drone image dataset \texttt{2Drone}.
We then tested the trackers on 2D synthetic sequences in \texttt{Syn3Drone} dataset as summarized in Tab.~\ref{tab:03}. 
The detector trained with \texttt{2Drone(on)} often failed to detect drone, due to the lack of training dataset.
On the other hand, the detector trained by \texttt{2Drone(on+aug)} improved drone detection rate significantly and helped the tracker operate properly.
In addition, it reduced the drone ID switch cases.
This results support that effectiveness of our datasets in training a reliable drone detector and tracker.
We anticipate further enhancements of the tracker by fine-tuning it using additional test domain images.

\begin{table}[t]
	\centering
	\small
	\setlength\tabcolsep{5pt}
	\renewcommand{\arraystretch}{1.2}
	\begin{tabular}{c||c|c|c|c|c}
		\hline\noalign{\hrule height 1pt}
		\rowcolor[HTML]{EFEFEF}
		Sequences & Metric & Width & Height & Diagonal & Ours \\
		\hline
		\multirow{2}{*}{\#01} & MAE & 4.20 & 32.38 & \textbf{3.84} & 4.53 \\
		\cline{2-6}
		& RMSE & 5.05 & 48.38 & \textbf{4.24} & 5.63 \\
		\hline
		\multirow{2}{*}{\#02} & MAE & 1.34 & 17.23 & 4.96 & \textbf{1.23} \\
		\cline{2-6}
		& RMSE & 1.76 & 27.33 & 6.94 & \textbf{1.72} \\
		\hline
		\multirow{2}{*}{\#03} & MAE & \textbf{1.15} & 27.78 & 5.05 & 2.16 \\
		\cline{2-6}
		& RMSE & \textbf{1.58} & 48.49 & 9.04 & 3.80 \\
		\hline
		\multirow{2}{*}{\#04} & MAE & 5.00 & 23.08 & \textbf{2.38} & 5.10 \\
		\cline{2-6}
		& RMSE & 8.07 & 33.02 & \textbf{3.80} & 8.26 \\
		\hline
		\multirow{2}{*}{\#05} & MAE & 23.54 & 31.86 & \textbf{20.27} & 25.57 \\
		\cline{2-6}
		& RMSE & 27.51 & 38.02 & \textbf{22.71} & 35.05 \\
		\hline
		\multirow{2}{*}{\#06} & MAE & \textbf{2.75} & 25.31 & 4.23 & 3.03 \\
		\cline{2-6}
		& RMSE & \textbf{3.73} & 37.74 & 5.26 & 4.24 \\
		\hline
		\multirow{2}{*}{\#07} & MAE & 5.34 & 11.13 & 8.60 & \textbf{5.00} \\
		\cline{2-6}
		& RMSE & 6.89 & 18.70 & 11.14 & \textbf{6.45} \\
		\hline
		\multirow{2}{*}{\#08} & MAE & 3.35 & 39.30 & 8.11 & \textbf{2.91} \\
		\cline{2-6}
		& RMSE & 5.24 & 71.53 & 12.80 & \textbf{4.28} \\
		\hline
		\multirow{2}{*}{\#09} & MAE & 9.78 & 38.94 & 7.16 & \textbf{4.10} \\
		\cline{2-6}
		& RMSE & 13.79 & 58.78 & 10.05 & \textbf{5.85} \\ \hline\noalign{\hrule height 1pt}
		\hline
		\multirow{2}{*}{Average} & MAE & 5.52 & 29.99 & 7.33 & \textbf{4.96} \\
		\cline{2-6}
		& RMSE & 7.46 & 48.91 & 10.24 & \textbf{6.99} \\ \hline\noalign{\hrule height 1pt}
	\end{tabular}
	\vspace{5pt}
	\caption{Performance comparison of 3D trajectory reconstructions}
	\vspace{0pt}
	\label{tab:04}
\end{table}

\begin{figure}[t]
	\centering
	\begin{minipage}[b]{0.326\linewidth}
		\centering
		\includegraphics[width=\linewidth]{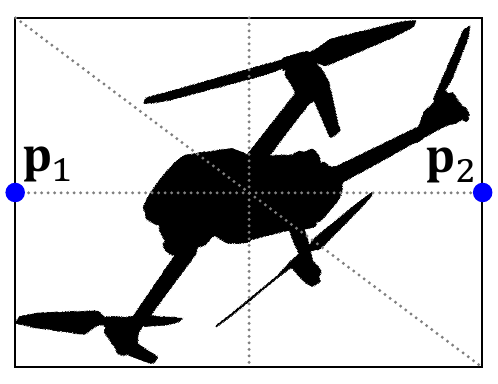}
		\caption*{Width-based}
	\end{minipage}
	\begin{minipage}[b]{0.326\linewidth}
		\centering
		\includegraphics[width=\linewidth]{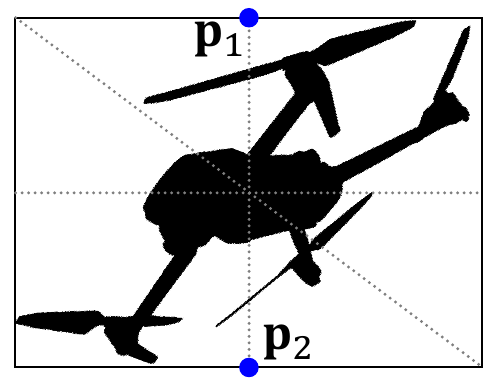}
		\caption*{Height-based}
	\end{minipage}
	\begin{minipage}[b]{0.326\linewidth}
		\centering
		\includegraphics[width=\linewidth]{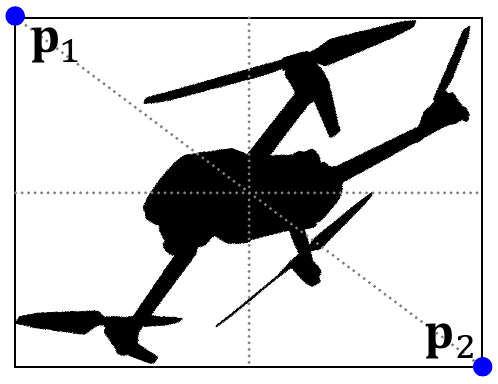}
		\caption*{Diagonal-based}
	\end{minipage}
	\caption{Determining drone 2D points $\mathbf{p}_1,\mathbf{p}_2$}
	\vspace{-10pt}
	\label{fig:8}
\end{figure}

\begin{figure*}[t]
    \centering
    \begin{subfigure}[t]{0.33\textwidth}
        \includegraphics[width=\linewidth]{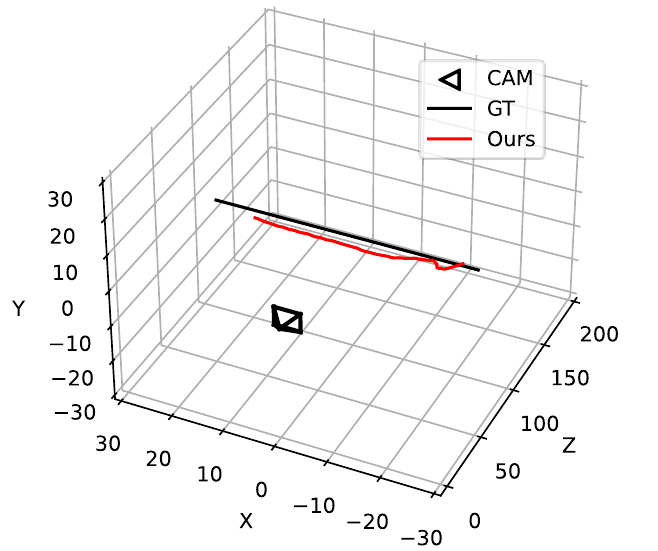}
        \caption{Sequence \#01}
    \end{subfigure}
    \hfill
    \begin{subfigure}[t]{0.33\textwidth}
        \includegraphics[width=\linewidth]{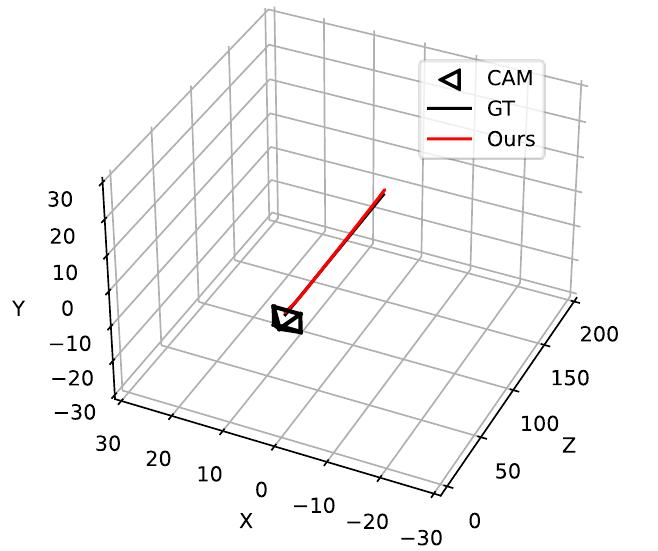}
        \caption{Sequence \#03}
        \vspace{5pt} 
    \end{subfigure}
    \hfill
    \begin{subfigure}[t]{0.33\textwidth}
        \includegraphics[width=\linewidth]{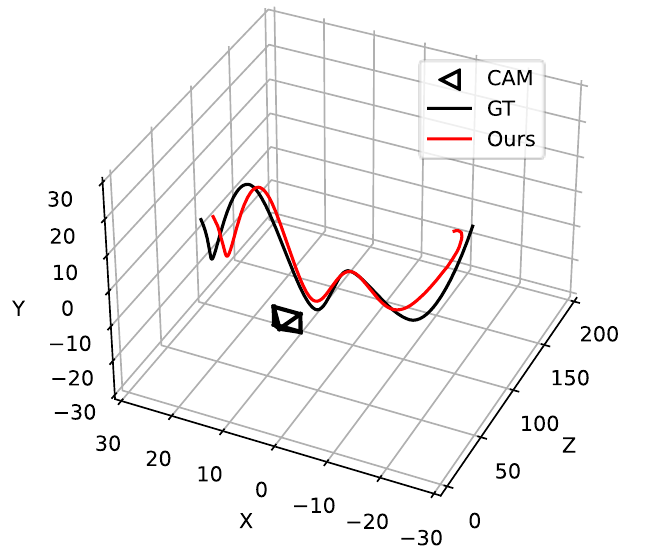}
        \caption{Sequence \#06}
        \vspace{5pt} 
    \end{subfigure}

    \vspace{10pt} 
    \begin{subfigure}[t]{0.33\textwidth}
        \includegraphics[width=\linewidth]{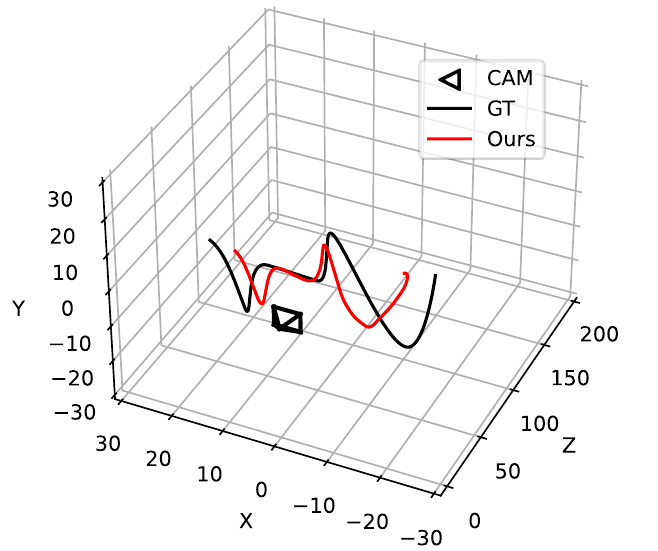}
        \caption{Sequence \#07}
        \vspace{5pt} 
    \end{subfigure}
    \hfill
    \begin{subfigure}[t]{0.33\textwidth}
        \includegraphics[width=\linewidth]{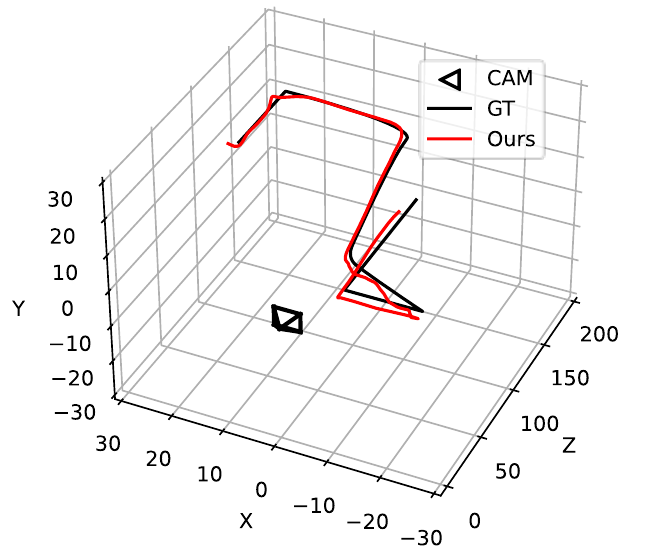}
        \caption{Sequence \#08}
    \end{subfigure}
    \hfill
    \begin{subfigure}[t]{0.33\textwidth}
        \includegraphics[width=\linewidth]{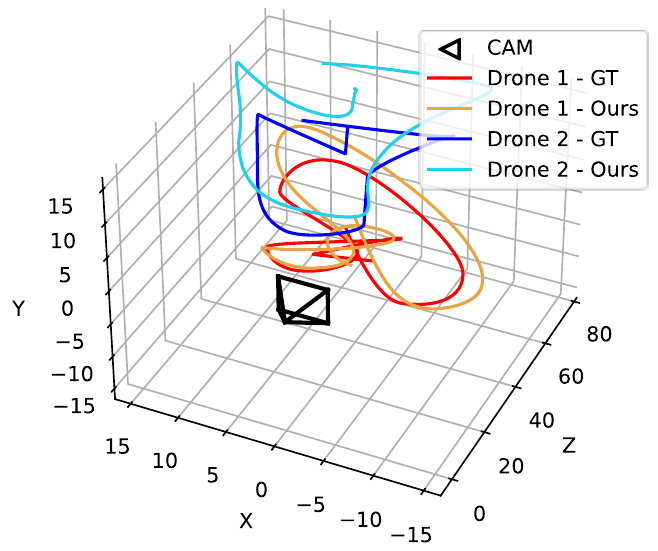}
        \caption{Sequence \#09}
        \vspace{5pt} 
    \end{subfigure}
    
    \caption{Qualitative results of 3D drone trajectory reconstruction based on the proposed methods}
    \label{fig:9}
\end{figure*}

\subsection{3D trajectory reconstruction results}

To validate the proposed methods, we employed the \texttt{Syn3Drone} dataset. During the experiments, we utilized ground-truth 2D drone bounding boxes to solely evaluate the effectiveness of our 3D trajectory reconstruction methods, independent of the performance of the drone detectors.
Initially reconstructed 3D trajectories of drones may contain error due to limited information of a single camera setting.
To mitigate reconstruction errors, we perform a moving average filter on the initial 3D trajectory. Figure.~\ref{fig:7} shows the trajectory smoothing results of Seq.~\#01 compared to the initial estimation. While a larger window size can provide a smoother trajectory, we empirically observed that window size of 5 yields the best 3D trajectory reconstruction performance.

To prove the effect of 2D drone rotation estimation proposed as in Fig.~\ref{fig:3}, we compared three simple approaches for determining 2D drone points $\mathbf{p}_1,\mathbf{p}_2$ as shown in Fig.~\ref{fig:8}. 
Table.~\ref{tab:04} presents 3D trajectory reconstruction performances of the different approaches and our method. 
The proposed method (ours) performed the best average reconstruction performance in terms of both evaluation metrics (MAE, RMSE). 
Especially, it showed superior performances in the most complex scenarios~(Seq.~\#08 and \#09) including drone rotation and non-linear motion.  
We can see qualitative trajectory reconstruction results of our methods in Fig.~\ref{fig:9}. 
Compared to the ground-truth trajectories, our methods performed reliable and reasonable reconstruction.
The results indicate that the proposed method efficiently handles the complex scenario, and demonstrate the potential for applying our methods to the common surveillance systems using a single static camera.

\section{Conclusions}
In this work, we proposed a novel framework for reconstructing 3D trajectories of drones using a single static camera. 
To this end, we exploited calibrated to leverage the relationship between 2D and 3D spaces, and tracked the drones in 2D images based on the drone tracker.
For the tracker we augmented the 2D drone image dataset, and trained an accurate 2D drone detector.
Furthermore, we proposed the 2D drone rotation estimation method.
By combining the 2D drone rotation information with its actual length, we geometrically inferred the 3D trajectories of drones in the camera coordinate system.
The experimental results showed that the proposed methods can perform reliable 3D drone trajectory reconstruction, and demonstrated
the potential for applying our framework in common surveillance systems using a single static camera.
Also, if the framework possesses pre-learned information for each class, it can be applied not only to drones but also to other objects that maintain a consistent shape and move freely. As a result, the system proposed in this study could find applications in diverse fields, including surveillance and security systems.

However, there exists a constraint when it comes to 2D rotation estimation for 3D objects such as drones. It is challenging to precisely estimate the absolute rotation of 3D objects solely from 2D images. The lack of depth information in 2D images makes accurately determining the rotation angle of a drone a complex issue. Therefore, the results of the proposed 2D rotation estimation may possess uncertainties.
To overcome these limitations, our future work will explore the integration of 6D pose estimation study~\cite{xiang2017posecnn,brachmann2014learning}. 6D pose estimation involves utilizing deep learning techniques to estimate the translation and rotation of objects covered in this study. By employing such approaches, if we achieve more accurate estimations of 3D objects, we can anticipate a more precise comprehension and analysis of drone trajectories. This enhancement, in turn, has the potential to elevate the accuracy of our framework and enable highly precise reconstructions of drone trajectories.



{\small
\bibliographystyle{ieee_fullname}
\bibliography{egbib}
}

\end{document}